\documentclass[journal]{IEEEtran}
\usepackage{graphicx}
\usepackage{amsfonts}

\usepackage[english]{babel}
\usepackage{amsmath, amssymb}
\usepackage{float}
\usepackage{makecell}
\usepackage{multirow}
\usepackage{cite}
\usepackage{amsmath,amssymb,amsfonts}
\usepackage{algorithmic}
\usepackage{graphicx}
\usepackage{textcomp}
\usepackage{xcolor}

\usepackage{subfigure}
\usepackage{epstopdf}

\usepackage{graphicx}
\usepackage{comment}
\usepackage{amsmath,amssymb} 
\usepackage{color}
\usepackage{enumerate}

\usepackage{booktabs}
\usepackage{threeparttable}

\usepackage{bbding}
\usepackage{float}

\usepackage{soul}
\soulregister\cite7 
\soulregister\citep7 
\soulregister\citet7 
\soulregister \ref7 
\soulregister\pageref7 

\ifCLASSINFOpdf
\else
\fi
\begin{document}
%
\title{Unsupervised Learning of 3D Scene Flow with 3D Odometry Assistance \thanks{*This work was supported in part by the Natural Science Foundation of China under Grant 62073222, Grant U21A20480, and Grant U1913204; in part by the Science and Technology Commission of Shanghai Municipality under Grant 21511101900; and in part by the Open Research Projects of Zhejiang Laboratory under Grant 2022NB0AB01. The first two authors contributed equally. Corresponding Author: Hesheng Wang.}
\thanks{G. Wang, Z. Feng, and H. Wang are with Department of Automation, Key Laboratory of System Control and Information Processing of Ministry of Education, Key Laboratory of Marine Intelligent Equipment and System of Ministry of Education, Shanghai Engineering Research Center of Intelligent Control and Management, Shanghai Jiao Tong University, Shanghai 200240, China.
}
\thanks{C. Jiang is with Engineering Research Center of Intelligent Control for Underground Space, Ministry of Education, School of Information and Control Engineering, Advanced Robotics Research Center, China University of Mining and Technology, Xuzhou 221116, China.}}
%
%
%

\author{Guangming~Wang,
       Zhiheng Feng, Chaokang Jiang, and Hesheng Wang}

\markboth{Journal of \LaTeX\ Class Files,~Vol.~14, No.~8, August~2015}%
{Shell \MakeLowercase{\textit{et al.}}: Bare Demo of IEEEtran.cls for IEEE Journals}
%



\maketitle

\begin{abstract}
Scene flow represents the 3D motion of each point in the scene, which explicitly describes the distance and the direction of each point's movement. Scene flow estimation is used in various applications such as autonomous driving fields, activity recognition, and virtual reality fields. As it is challenging to annotate scene flow with ground truth for real-world data, this leaves no real-world dataset available to provide a large amount of data with ground truth for scene flow estimation. Therefore, many works use synthesized data to pre-train their network and real-world LiDAR data to finetune. Unlike the previous unsupervised learning of scene flow in point clouds, we propose to use odometry information to assist the unsupervised learning of scene flow and use real-world LiDAR data to train our network. Supervised odometry provides more accurate shared cost volume for scene flow. In addition, the proposed network has mask-weighted warp  layers to get a more accurate predicted point cloud. The warp operation means applying an estimated pose transformation or scene flow to a source point cloud to obtain a predicted point cloud and is the key to refining scene flow from coarse to fine. When performing warp operations, the points in different states use different weights for the pose transformation and scene flow transformation. We classify the states of points as static, dynamic, and occluded, where the static masks are used to divide static and dynamic points, and the occlusion masks are used to divide occluded points. The mask-weighted warp layer indicates that static masks and occlusion masks are used as weights when performing warp operations. Our designs are proved to be effective in ablation experiments. The experiment results show the promising prospect of an odometry-assisted unsupervised learning method for 3D scene flow in real-world data. 
\end{abstract}


\IEEEpeerreviewmaketitle

\section{Introduction}
%
%
%
%


\IEEEPARstart{M}{otion} information at each point in 3D space is essential for dynamic scene perception and reconstruction in the field of robotics and autonomous driving \cite{behl2017bounding}. There are many practical applications of scene flow, such as object detection \cite{behl2017bounding, cao2019learning}, multi-object tracking \cite{wang2020pointtracknet, wang2022interactive}, and segmentation \cite{behl2017bounding, liu2019meteornet, lv2018learning, ma2019deep}. 3D scene flow consists of a motion vector for each point in the consecutive frames of point clouds. It is difficult for common sensors such as LiDAR and RGBD cameras to collect this motion information directly, which elicits the task of scene flow estimation. It is difficult to annotate the ground truth for 3D scene flow from real-world LiDAR data. Therefore, many works\cite{HALFlow,Hplflownet,Flownet3d,Pointpwc} use unsupervised learning for scene flow estimation. Many deep learning-based works of scene flow estimation use synthesized data such as the FlyingThings3D dataset \cite{Flyingthing3d} to pre-train their network. However, scene flow estimation methods based on synthesized datasets have obvious drawbacks. There are significant differences between the synthesized dataset and the real-world dataset, which results in models trained on the synthesized dataset not being well adapted to the real-world scenes \cite{Pointpwc}. Because a large amount of 3D point clouds can be easily accessed due to the popularity of LiDAR sensors, unsupervised learning of 3D scene flow in real-world datasets shows great potential.

Some works \cite{Justgowith,Pointpwc,sfgan} introduce a variety of unsupervised loss functions at the beginning of unsupervised learning of 3D scene flow development. They calculate the loss based on the predicted point cloud of the second frame and the real point cloud of the second frame, where the predicted point cloud is generated by summing the predicted scene flow and the point cloud of the first frame. The computed loss is back-propagated to the scene flow estimation network. These methods do not change the internal structure of the network, and they just 
estimate the scene flow as each point motion vector directly without classifying the point motion vectors. Tishchenko et al. \cite{Ego-Motion} attempt to distinguish between moving and static points in the scene by introducing a two-stage network. They use an existing scene flow network to estimate the residual flow between the point cloud of the second frame and the point cloud transformed by ego-motion estimation. The distinction of point states at the level of network prediction results does not essentially distinguish the motion vectors of different states and suffers from a high error rate in the real scene. SLIM \cite{slim} generates an unsupervised motion segmentation signal based on the difference between the rigid ego-motion estimation and the original scene flow prediction. However, their network predicts the scene flow based on the RAFT \cite{raft} which is an optical flow network based on bird's eye view of point cloud. 3D geometric information may be lost during the inference. The common occlusion problem in the real world is also ignored.

To overcome the above challenges, we propose the LiDAR odometry-assisted unsupervised learning network for 3D scene flow. Compared to 3D scene flow ground truth, the odometry ground truth can be easily obtained through sensors such as Global Positioning System (GPS) and Real-time Kinematic (RTK). The purpose of this paper is to utilize inexpensive apriori knowledge to guide the learning of complex knowledge. Both the scene flow network and the odometry network contain Pyramid, Warping, and Cost volume (PWC) structure. Among them, accurate warp in the refinement process from coarse to fine is essential to construct cost volume to perceive point motion.  We classify the points as static, dynamic, and occluded. The points in different states use different weights for the pose information and scene flow information to get more accurate predicted point clouds when performing warp operations. Because odometry network is end-to-end supervised learning, better odometry estimation performance makes it better to assist 3D scene flow estimation. 

In summary, the main contributions of this paper are as follows:
\begin{itemize}
\item[$\bullet$] The novel unsupervised learning network of 3D scene flow with odometry assistance is proposed. We use the prior pose information which can be measured by sensors to assist the learning of scene flow which can not be measured by sensors. The shared cost volume provide by the odometry is proposed to calculate the occlusion perception cost volume of scene flow. Because the odometry is trained in a supervised way, the cost volume of odometry has better ability to perceive the motion of consecutive point clouds.

\item[$\bullet$] This paper proposes a new divide-and-conquer point cloud warp layer that makes points in the different states use different weights for odometry and scene flow information to get their predicted coordinates. For the warp operation, the pose transformation is more accurate for static and non-occluded points, while the scene flow is more accurate for dynamic points. Therefore, the warp operation gives a more accurate predicted point cloud.

\item[$\bullet$] 
 The evaluation results of our method trained on real-world LiDAR point clouds show the promising learning capability in the real-world scene. Ours without any fine-tuning outperforms PointPWC-Net \cite{Pointpwc}. The experiment results of ablation studies show the indispensable role of static mask and occlusion mask and the effectiveness of pose information in the point cloud warp layer.

\end{itemize}

Our paper is organized as follows: Section \ref{sec:related_work} is about some related works. Section \ref{sec:approach} introduces our network and the module details in the network. Training loss is in Section \ref{sec:loss_function}. The training details, training dataset, evaluation dataset, experiment results, and ablation studies are in Section \ref{sec:experiment}. The Section \ref{sec:conclusion} is on our conclusion.

\section{Related Work} \label{sec:related_work}

\subsection{Deep Learning for 3D Scene Flow}

Many works \cite{liu2019flownet3d,flot,HALFlow,Flownet3d,Hplflownet,residual} on 3D scene flow learning mainly focus on supervised learning. FlowNet3D \cite{liu2019flownet3d} is the first to propose the learning of 3D scene flow from point cloud pairs with an end-to-end approach. FlowNet3D \cite{liu2019flownet3d} associates learned point space locality and geometric similarity by introducing a flow embedding layer. FLOT \cite{flot} introduces optimal transport to the task of 3D scene flow estimation to constrain the search for point matching according to the graph matching. PointPWC-Net \cite{Pointpwc} introduces point cloud cost volume, up-sampling and warping layers to perceive the motion between two consecutive frames of point clouds. The weight in cost volume proposed by PointPWC-Net  \cite{Pointpwc} is decided only by Euclidean space, Wang et al. \cite{HALFlow} develop the attentive cost volume considering feature space and propose a hierarchical attention learning network for scene flow estimation with two different attentions. These methods rely excessively on synthesized datasets. Gojcic et al. \cite{Gojcic_2021_CVPR} consider scene flow in combination with other 3D tasks. They infer rigid scene flow at the object level rather than at the point level. The annotations generated from background segmentation and ego-motion weakens the limitations of dense scene flow supervision.

Because it is difficult to obtain the ground truth for 3D scene flow from real-world LiDAR data, unsupervised learning of 3D scene flow on point clouds is a research field. PointPWC-Net \cite{Pointpwc} introduces three loss functions, Chamfer loss, smoothness constraint, and Laplacian regularization to achieve unsupervised learning of 3D scene flow without any annotation. Mittal et al. \cite{Justgowith} propose cycle consistency loss to constrain the time consistency of the predicted scene flow. They also propose nearest neighbor loss, where the nearest neighbor of the predicted point cloud found in the second frame of the point cloud is considered a pseudo-ground truth. To distinguish between ego-motion and object motion, Tishchenko et al. \cite{Ego-Motion} propose to learn rigid transformations in a pair of point clouds using a pose estimation network. The non-rigid residual flow is learned by the scene flow estimation network. SLIM \cite{slim} adapts the RAFT \cite{raft} network structure to iteratively update two newly designed logits in addition to the output prediction flow. The first logit classifies the points as static or moving. The second logit characterizes the confidence level of the output flow. OGSF \cite{occlusion} presents the correlation layer that simultaneously learns the scene flow and the occlusion mask. This mask-weighted cost volume weakens the effect of the occlusion on the 3D scene flow estimation. To resolve the poor model adaptation caused by the large differences between synthetic datasets and real-world scenes, Jin et al. \cite{jin2022deformation} develop a large-scale synthetic scene flow dataset and propose a mean-teacher-based domain adaptation framework. When searching for correspondence points in scene flow estimation, distant points are ignored, but these distant points may be the actual matching ones. To solve the problem of ignoring distant points, Wang et al. \cite{wang2022matters} propose an all-to-all flow embedding layer with backward reliability validation in the initial estimation module of 3D scene flow.
\begin{figure*}[ht]
	\centering
	\includegraphics[width=\textwidth]{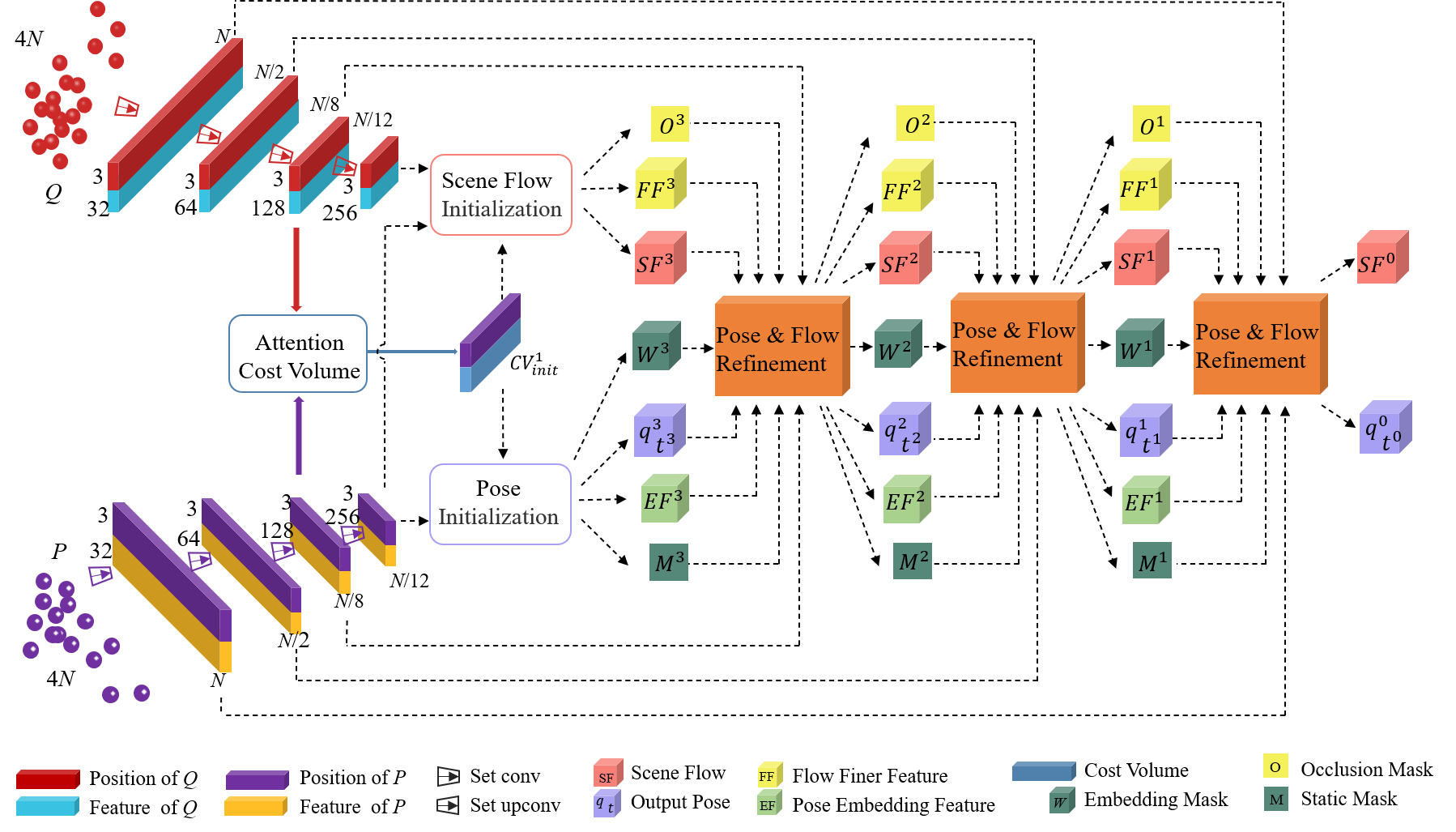}
	\caption{\textbf{LiDAR odometry-assisted unsupervised learning network structure for 3D scene flow.} The two frames of point clouds are first sampled through hierarchical point feature encoding introduced in Section \ref{sec:fearure_ab}, and the initial cost volume $CV_{init}^1$ is generated using attentive cost volume introduced in Section \ref{sec:Occ_cost}.1. Then, they are input into the scene flow initialization module introduced in Section \ref{sec:init}.1 and the pose initialization module introduced in Section \ref{sec:init}.2, respectively. Coarse scene flow ${SF}^3$, occlusion mask ${O}^3$, quaternion ${q}^3$, translation vector ${t}^3$, static mask ${M}^3$, embedding mask ${W}^3$, finer flow feature ${FF}^3$, and pose embedding features ${EF}^3$ are generated through the two initialization modules. Next, they are input into the pose and flow refinement module introduced in Section \ref{sec:refine_layer}, and the final refined scene flow estimation ${SF}^0$, quaternion ${q}^0$, and translation vector ${t}^0$ are output after layer-by-layer refinement.}
	\label{fig:network}
\end{figure*}
\subsection{Deep Learning for LiDAR Odometry} 
In the beginning, many works \cite{Lodonet,wang2019deeppco,li2019net,velas2018cnn} project 3D point clouds of consecutive frames onto 2D images and learn odometry using 2D processing methods to reduce the learning difficulty from sparse and unstructured point clouds. As the development of 3D deep learning, it becomes a promising direction that inferring the 6-DOF pose directly from 3D LiDAR point clouds. PWCLO-Net \cite{PWCLO} constructs a deep network for learning pose directly from 3D LiDAR point clouds in a coarse-to-fine method. Furthermore, they propose a trainable mask to perceive the motion state of each point. Wang et al. \cite{efficient} design the projection-aware set-conv layer, the projection-aware cost volume module, and the projection-aware set-upconv layer, which achieve efficient LiDAR odometry estimation. StickyPillars \cite{stickypillars} performs contextual aggregation of sparse 3D points by graph neural networks. They estimate odometry by explicit 3D feature correspondence.

\section{Unsupervised Learning of 3D Scene Flow with 3D Odometry Assistance} \label{sec:approach}
\subsection{Network Architecture}
As shown in Fig. \ref{fig:network}, point clouds $P=\{p_i|p_i \in \mathbb{R}^3 \}_{i=1}^{4N}$ and  $Q=\{q_i|q_i \in \mathbb{R}^3 \}_{i=1}^{4N}$ from consecutive frames are passed into the network. Firstly, encoded features are extracted from these two frames of point clouds through the siamese pyramid \cite{chopra2005learning} introduced in Section \ref{sec:fearure_ab}. Then, feature matching is performed through the attentive cost volume \cite{HALFlow} introduced in Section \ref{sec:Occ_cost}.1 to generate initial cost volume $CV_{init}^1$. 

Next, the initial cost volume and the encoded features at the last level of $Q$ and $P$ are input into the scene flow initialization module introduced in Section \ref{sec:init}.1 to produce coarse scene flow information. In the scene flow initialization module, a module named occlusion perception cost volume introduced in Section \ref{sec:Occ_cost} is used to generate cost volume with occlusion perception. The scene flow information includes scene flow ${SF}$, occlusion mask ${O}$, and finer flow feature ${FF}$. In the same way, the initial cost volume $CV_{init}^1$ and the encoded features at the last level of $P$ are input into the pose initialization module introduced in Section \ref{sec:init}.2 to produce coarse pose information. The pose information includes quaternion ${q}$, translation vector ${t}$, static mask ${M}$, embedding mask $W$, and pose embedding features ${EF}$.

Then, a refinement module called pose and scene flow refinement is applied to refine pose and scene flow estimation, which is introduced in Section  \ref{sec:refine_layer}. Finally, the network outputs the estimated pose, scene flow, static mask, and occlusion mask for each level.   

\subsection{Hierarchical Point Feature Encoding} \label{sec:fearure_ab}

The collected LiDAR point clouds are usually disorganized and sparse in 3D space. 
The features of the point cloud are extracted by encoding the coordinate of points in the point cloud. The hierarchical siamese feature pyramid structure is used for the feature extraction of the point cloud. The feature dimension of the point cloud increases as the pyramid level rises.

Farthest point sampling (FPS) \cite{qi2017pointnet++} is used to sample point cloud from ($l$+1)-th level to  $l$-th level, and $K$ nearest neighbors (KNN) is used to select $K$ nearest points for each sampled  point. 
The formula for calculating the features of point cloud at the ($l$+1)-th level from the features of point cloud at the $l$-th level is as follows:
\begin{equation}
f'_i =  \mathop {{\text{MAX}}}\limits_{k = 1,...,K}(\text{MLP} ((p_i^k-p_i) \oplus f_i^k \oplus f_i),
\end{equation}
where $p_i \in \mathbb{R}^3$ is the coordinate of the $i$-th point in point cloud $P$ and $f_i$ is the feature of $p_i$. $p_i^k$ is the coordinate of the $k$-th  point in the $K$ nearest neighbors of $p_i$ and $f_i^k$ is the feature of point $p_i^k$. $f'_i$ is the feature of point cloud at the ($l$+1)-th level. $\oplus$ means the concatenation of two vectors. The $\mathop{\text{MAX}}\limits_{k = 1,...,K}$ means the max-pooling operation and MLP is Multi-Layer Perceptron (MLP). The initial features of the point cloud are its 3D coordinates.

\subsection{Occlusion Perception Cost Volume} \label{sec:Occ_cost}
\begin{figure}[t]
\centering
\includegraphics[width=0.48\textwidth]{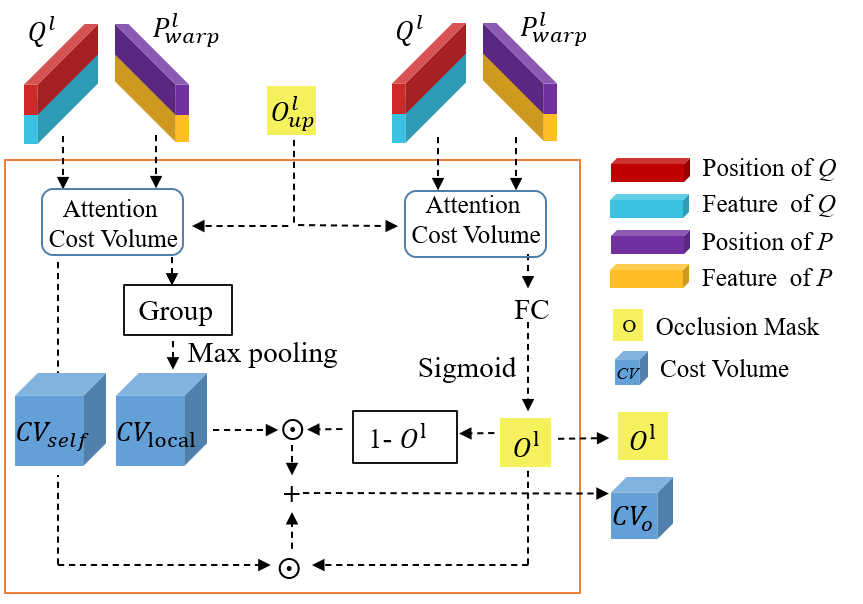}
\caption{\textbf{The details of occlusion perception cost volume.} Attentive cost volume introduced in Section \ref{sec:Occ_cost}.1 is used to get cost volume without occlusion perception. The two attentive cost volume shown in the figure are the same structure, except that the number of Multi-Layer Perceptron (MLP) layers in them is different. Group means getting $K$ nearest neighbors of a point.} \label{fig:cost}
\end{figure}

The cost volume module is used to calculate the feature matching cost of two consecutive frames of point clouds after obtaining their encoded features. The cost volume indicates the matching information between the two frames of point clouds. In the real world, due to the influence of occlusion, the occluded points in $P$ are lost in $Q$. Such points are not suitable for calculating the cost volume. Therefore, the cost volume with an occlusion perception module similar to that proposed by OGSF \cite{occlusion} is introduced to get accurate cost volume. Different from \cite{occlusion}, we introduce the attentive cost volume into our module.

\subsubsection{Attention Cost Volume} \label{sec:attention}

Attentive cost volume is introduced in \cite{HALFlow} for calculating the cost volume of two point clouds. The cost volume $CV_{self}(p_i)$ obtained by this method is without occlusion perception between point cloud $P$ and point cloud $Q$. 

First, the position information between $p_i$ and its $K_1$ nearest neighbors $q_k$ ($k=1,2,...,K_1$) in point cloud $Q$ is encoded by the Eq.   (\ref{eq:pos_p2q}):
\begin{equation}
pos_{i,k} =p_i \oplus q_k \oplus (q_k-p_i)\oplus ||q_k-p_i||_2^2, \label{eq:pos_p2q}
\end{equation}
Then, the encoded feature $f_{q2p}(p_{i,k})$ between $p_i$ and its $K_1$ nearest neighbors $q_k$ ($k=1,2,...,K_1$) is calculated by the Eq.  (\ref{eq:f_p2q}):
\begin{equation}
f_{q2p}(p_{i,k}) = MLP(pos_{i,k} \oplus f_{p_i} \oplus f_{q_k}), \label{eq:f_p2q}
\end{equation}
where $f_{p_i}$ is the encoded feature of $p_i$, and $f_{q_k}$ is the encoded feature of $q_k$. For point $p_i$, when calculating the information of feature matching, both the distance and the feature similarity between $p_i$ and its $K_1$ nearest neighbors have  influence on the aggregation weights. Therefore, the weights  $w_{q2p}(i,k)$ for $f_{q2p}(p_{i,k})$ is calculated as follows:
\begin{equation}
w_{q2p}(i,k) = softmax(MLP(pos_{i,k} \oplus f_{q2p}(p_{i,k})). \label{eq:w_p2q}
\end{equation}
The cost volume $cost_{self}(p_i)$ between $p_i$ and its $K_1$ nearest neighbors in point cloud $Q$ is represented as:
\begin{equation}
cost_{q2p}(p_i) = \sum_{k=1}^{K_1} f_{q2p}(p_{i,k}) \odot w_{q2p}(i,k),
\end{equation}
where $\odot$ means dot product. $p_k$ is the $k$-th point in the $K_1$ nearest neighbors of $p_i$. 

Next, the feature aggregation in point cloud $P$ is performed to expand the received fields of the cost volume \cite{hu2020randla}. The weights $w_{p2p}(i,k)$ of $p_i$ and its $K_2$ nearest neighbors $p_k$ ($k=1,2,...,K_2$) in $P$ is calculated by using the similar method shown as Eq. (\ref{eq:w_p2q}). Finally, the attentive cost volume $CV_{self}(p_i)$ is calculated by the formula as follows:
\begin{equation}
CV_{self}(p_i) = \sum_{k=1}^{K_2} cost_{q2p}(p_{k}) \odot w_{p2p}(i,k). \label{eq:CVself}
\end{equation}

\subsubsection{Occlusion Perception} \label{sec:perception}

The feature $f_O$ of the occlusion mask is calculated by using the attentive cost volume. $f_O$ is passed through an FC layer and sigmoid activation function to obtain the occlusion mask $O$. The mask is a weight matrix whose element value is from 0 to 1. The value represents the probability that the point is a non-occluded point. The formula is:
\begin{equation}
   O=sigmoid(FC(f_O)). \label{eq:occlusion}
\end{equation}
 
When a point $p_i$ is occluded, its cost volume $CV_{self}(p_i)$ is unreliable. The cost volume of this point should be guided by its $K$ nearest neighbors. Therefore, the final cost volume $CV_o(p_i)$ with occlusion perception is designed as:
\begin{equation}
     CV_{local}(p_i) = \mathop {{\text{MAX}}}\limits_{s_k\in N(p_i)}(CV_{self}(s_k)),
\end{equation}
\begin{equation}
CV_o(p_i) =  O(p_i)CV_{self}(p_i)+ (1-O(p_i))CV_{local}(p_i), \label{eq:cost_CV}
\end{equation}
where the $N(p_i)$ are the $K$ nearest neighbors of point $p_i$. $O(p_i)$ is the occlusion mask value of $p_i$.  $CV_{local}(p_i)$ is the cost volume guided by $N(p_i)$. As shown in Eq.  (\ref{eq:cost_CV}), we focus on that the $CV_{self}(p_i)$ has a higher contribution for non-occluded points,  while the $CV_{local}(p_i)$ has a higher contribution for occluded points.  

\subsection{Scene Flow Initialization and Pose Initialization} \label{sec:init}

The scene flow initialization module and pose initialization module are used to generate initial coarse estimation of scene flow and pose. 

\subsubsection{Scene Flow Initialization}
The initial cost volume  $CV_{init}^1$ is sampled to get ${CV}_{init}^2$, ${CV}_{init}^2$ is sampled to get ${CV}_{init}^3$, and the ${CV}_{init}^3$ is up-sampled to get ${CV}_{init}^4$. An FC layer is used to generate ${SF}_{C}^3$ which is the initial estimation of the scene flow. The formula is:
\begin{equation}
    SF_{C}^3 = FC(CV_{init}^4).
\end{equation}
The predicted point cloud $P_{w}$ is obtained by adding ${SF}_{C}^3$ to the point cloud $P$. The point cloud $Q$, the features of $Q$, $P_{w}$, and ${CV}_{init}^2$ are input into the  occlusion perception cost volume to obtain the coarse estimated mask $O^3$ and the cost volume $CV$. The MLP is used to generate the coarse flow feature ${FF}^3$.  The formula is:
\begin{equation}
    FF^3 = MLP(CV \oplus CV_{init}^4  \oplus O^3).
\end{equation}

Finally, the output coarse scene flow ${SF}^3$ is:
\begin{equation}
    SF^3 = SF_{C}^3 + FC(FF^3).
\end{equation}

\begin{figure}[t]
\centering
\includegraphics[width=\linewidth]{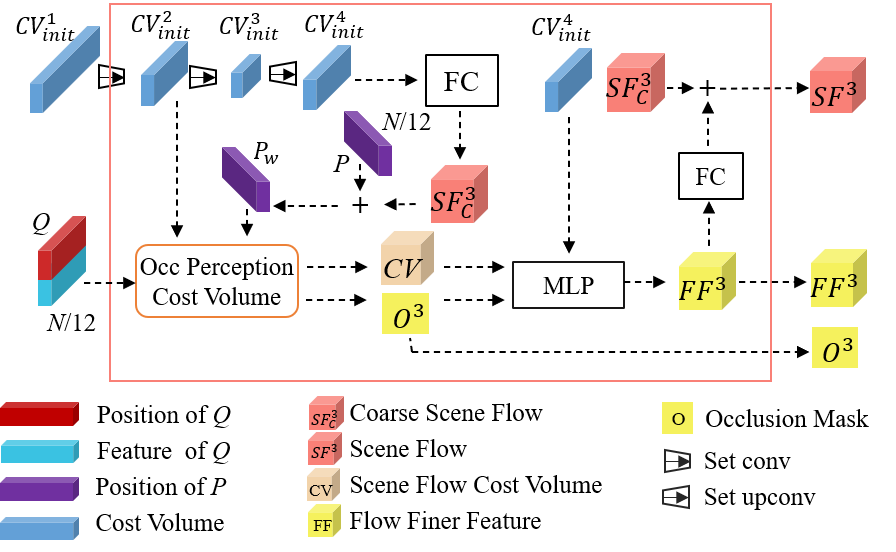}
\caption{\textbf{The details of scene flow initialization.} ``Occ perception cost volume" is introduced in Section \ref{sec:Occ_cost}. The module outputs initial scene flow ${SF}^3$, occlusion mask ${O}^3$, and finer flow feature ${FF}^3$.} \label{fig:sceneInit}
\end{figure}

\subsubsection{Pose Initialization}
The initial cost volume $CV_{init}^1$ is sampled to get  $CV_{init,p}^2=\{cv_{i,p}^2 | cv_{i,p}^2 \in \mathbb{R}^c \}$. The MLP is used to generate the embedding feature ${EF}^3$. For the odometry task, ego-motion is affected by moving objects in the real world. PWCLO-Net \cite{PWCLO} proposes a embedding mask $W$ to weight the embedding features of dynamic and static points. Because the $w^3$ is on the feature dimension of the embedding feature, an FC layer is used to obtain the static mask $M^3$. The formula is:
\begin{gather}
 M^3 = FC(W^3).
\end{gather}
We use $W^3$ as the weight to calculate the $feat_{qt}$ which is the feature that contains information about the pose transformation. The formual is: 
\begin{equation} 
 feat_{qt}=dropout(FC(\sum_{i=1}^{N^4} w_i \odot cv_{i,p}^2)).\label{formula:feat}
\end{equation}

Finally, the coarse estimation of quaternion $q^3$ and translation vector $t^3$ is:
\begin{equation} 
\begin{gathered} 
q^3=\frac{FC(feat_{qt})}{|FC(feat_{qt})|},\\
t^3=FC(feat_{qt}). \label{formula:qt}
\end{gathered} 
\end{equation}

\begin{figure}[t]
\centering
\includegraphics[width=0.9\linewidth]{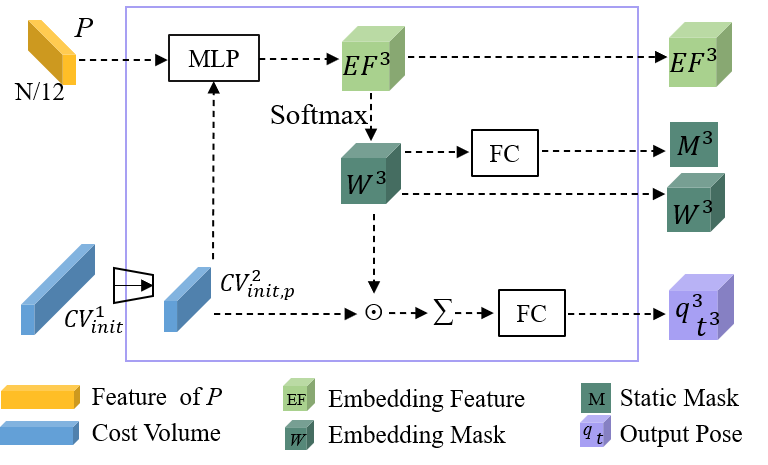}
\caption{\textbf{The details of pose initialization.} The module outputs initial quaternion ${q}^3$, translation vector ${t}^3$, static mask ${M}^3$, embedding mask $W^3$, and pose embedding features ${EF}^3$.} \label{fig:poseInit}
\end{figure}
\subsection{Hierarchical Pose and Scene Flow Refinement}
\label{sec:refine_layer}
\begin{figure*}[t]
\centering
\includegraphics[width=0.85\textwidth]{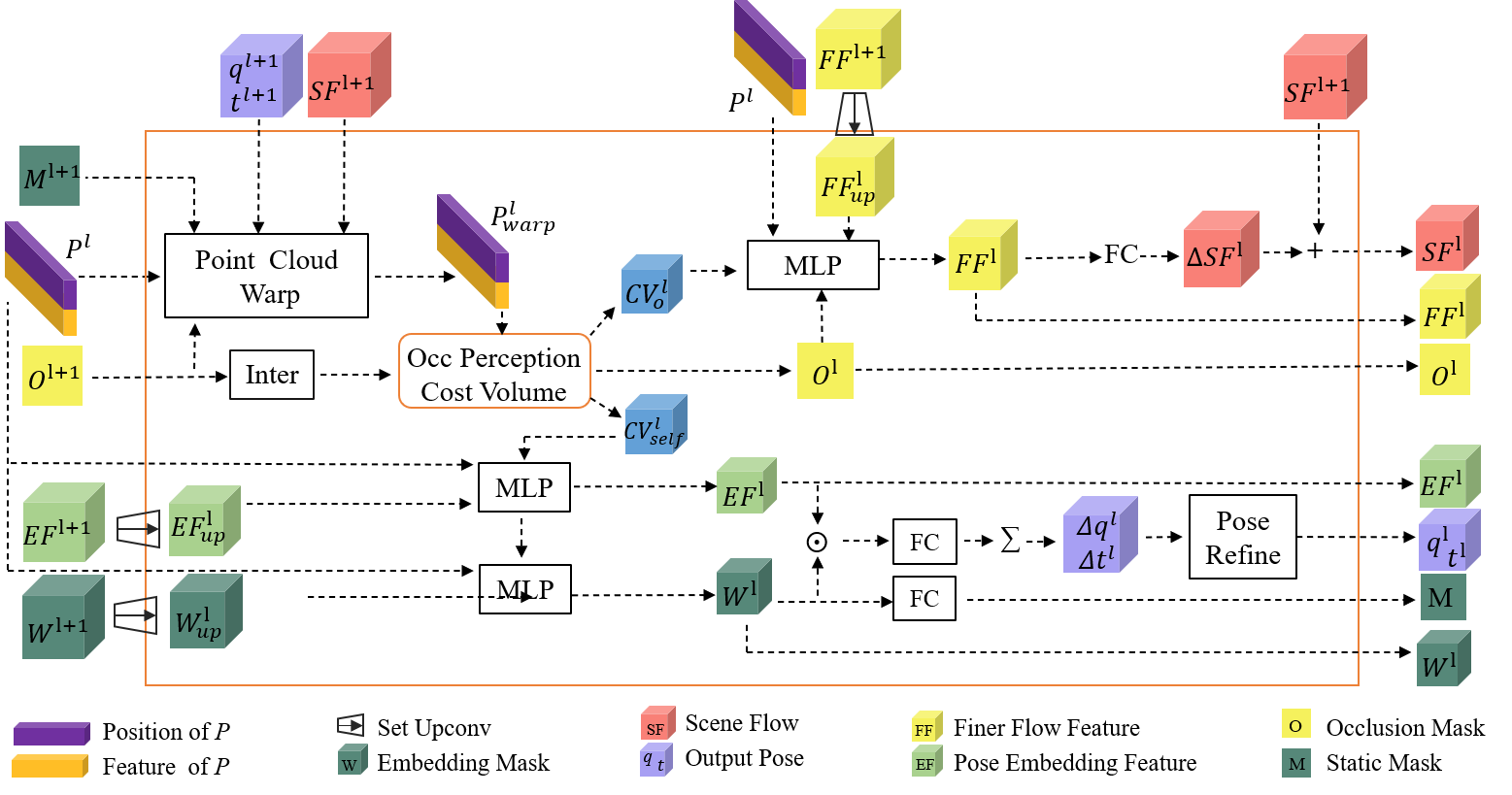}
 \caption{\textbf{The details of pose and flow refinement.} The module is applied to refine pose and scene flow estimation and finally outputs the estimated pose, scene flow, static mask, and occlusion mask for each level. The module includes point cloud warp layer, set upconv layer, pose estimation, and scene flow estimation. Pose refinement represents the operation as shown in Eq. (\ref{eq:pose_t}).}
  \label{fig:refine} 
\end{figure*}
\begin{figure}[t]
\centering
\includegraphics[width=0.48\textwidth]{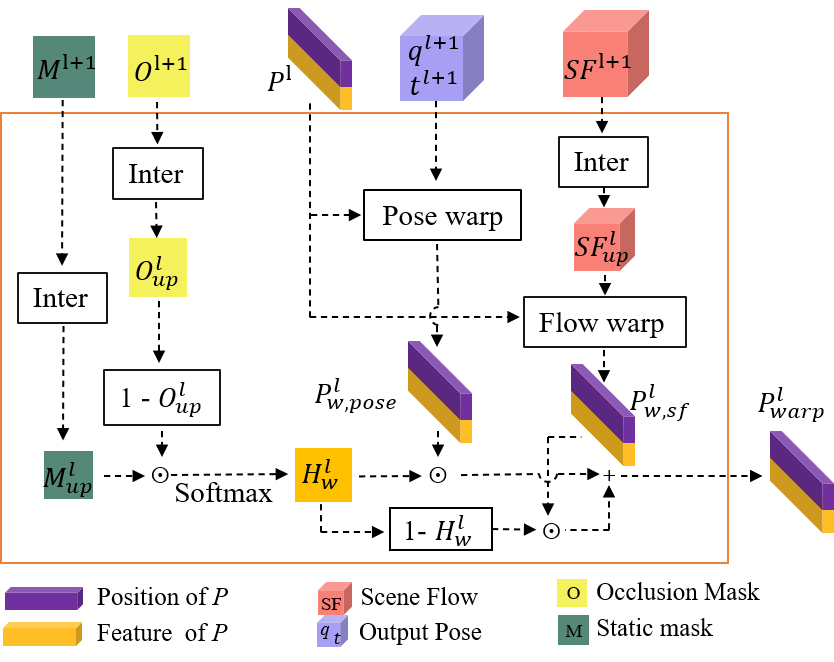}
  \caption{\textbf{The Detail of the point cloud warp layer with occlusion mask and static mask.} ``Inter" means interpolation up-sampling. Pose warp and flow warp are the warp operation. $H_w^l$ is obtained from the dot product result of two masks. $H_w^l$ denotes the weight of pose in warp operation. 1-$H_w^l$ denotes the weight of scene flow in warp operation.}
  \label{fig:warp} 
\end{figure}

The joint refinement process of scene flow and pose is proposed as shown in Fig.  \ref{fig:refine}.

\subsubsection{Set Upconv Layer} \label{sec:set_up}
Set upconv layer is adapted to propagate embedding features $EF^{l+1} = \{ef_i| ef_i \in \mathbb{R}^{c}\}_{i=1}^{N^{l+1}}$, flow feature $FF^{l+1} = \{ff_i| ff_i \in \mathbb{R}^{c}\}_{i=1}^{N^{l+1}}$, and embedding  mask $W^{l+1} = \{w_i| w_i \in \mathbb{R}^{c}\}_{i=1}^{N^{l+1}}$ from ($l$+1)-th level to $l$-th level. $EF^{l}_{up} =\{ef_i| ef_i \in \mathbb{R}^{c}\}_{i=1}^{N^l}$, flow feature $FF^{l}_{up} = \{ff_i| ff_i \in \mathbb{R}^{c}\}_{i=1}^{N^l}$, and embedding mask $W^{l}_{up} = \{w_i| w_i \in \mathbb{R}^{c}\}_{i=1}^{N^l}$ are obtained by the set upconv layer.

Because the number of points in the ($l$+1)-th level is less than that in the $l$-th level. The interpolation up-sampling method based on three nearest neighbors (Three-NN) \cite{qi2017pointnet++} is used to up-sample the static mask $M^{l+1}$, occlusion mask $O^{l+1}$, and scene flow $SF^{l+1}$ to obtain the mask $M^l_{up}$, $O^l_{up}$, and $SF^l_{up}$ of the $l$-th level. 

\subsubsection{Point Cloud Warp Layer with Occlusion Mask and Static Mask} \label{sec:warp_layer} 
In the classical coarse-to-fine scene flow estimation method, the warp operation is the key to scene flow refinement. To get more accurate predicited point clouds, the point cloud warping layer with occlusion and static mask shown in Fig.  \ref{fig:warp} is designed. The warp operation of point clouds means that scene flow or pose estimated at the ($l$+1)-th level are applied to warp source point cloud to generate predicted point cloud $P^l_{warp}$. $P^l_{w,pose}$ and $P^l_{w,sf}$ denote predicted point clouds obtained using pose estimation and scene flow estimation, respectively. The formula to calculate $P^l_{w,pose}$ and $P^l_{w,sf}$ is:
\begin{equation}
[0,p_{pose,i}^l]=q^{l+1}[0,p_i^l](q^{l+1})^{-1}+[0,t^{l+1}], \label{eq:warp_pose}
\end{equation}
\begin{equation}
 p_{sf,i}^l=p_i^l+sf^{l}_{i,up}, \label{eq:warp_sf}
\end{equation}
where Eq. (\ref{eq:warp_pose}) means quaternion $q^{l+1}$ and translation vector $t^{l+1}$ are applied to warp $P^l$ to generate $P^l_{w,pose} = \{p^l_{pose,i} | p^l_{pose,i} \in  \mathbb{R}^3 \}$.  Eq. (\ref{eq:warp_sf}) means $SF^l_{up} = \{sf^l_{i,up}|sf^l_{i,up} \in \mathbb{R}^3 \}$ is applied to warp $P^l$ to generate $P^l_{w,sf} = \{p^l_{sf,i} | p^l_{sf,i} \in \mathbb{R}^3 \}$.

Static points are more accurate for pose estimation, while dynamic points or non-occluded points are more accurate for scene flow estimation. The mask $H^l_w = \{ h^l_i |h^l_i \in \mathbb{R}^1 \}$ used to calculate the weights of $P^l_{w,pose}$ and $P^l_{w,sf}$ is calculated as follows:
\begin{equation}
   H_w^l=softmax(M^l_{up} \odot (1-O^l_{up})). \label{eq:h_weight}
\end{equation}
Finally, the predicted point cloud $P_{warp}^l$ at the $l$-th level is represented as:
\begin{equation}
  P_{warp}^l = H_w^l \odot P^l_{w,pose} + (1-H_w^l) \odot  P^l_{w,sf}. \label{eq:warp_psf}
\end{equation}

\subsubsection{Pose Estimation} \label{sec:pose_est}
$EF^{l}_{up}$, $CV_{self}$, and the feature of $P^l$ are concatenated and input into an MLP to produce refined $EF^{l}$. Then, $M^{l}_{up}$, $EF^{l}$, and the feature of $P^l$ are concatenated and input into an MLP to produce refined $W^{l}$. The $feat_{qt}^l$ which contains the residual information of pose transformation can be obtained from embedding feature $EF^{l}$ and embedding mask $W^{l}$ using the Eq. (\ref{formula:feat}).

Then, the residual values $\Delta q$ and $ \Delta t$  can be obtained from $feat_{qt}^l$ using the Eq. (\ref{formula:qt}). In the pose estimation at the $l$-th level, the formula for calculating the refined $q^l$ and $t^l$ is:
\begin{equation} 
\begin{gathered} 
 q^l=q^{l+1} \Delta q^l,  \\
  [0,t^l]=\Delta q^l[0,t^{l+1}] (\Delta q^l)^{-1} + [0,\Delta t^l]. \label{eq:pose_t}
\end{gathered} 
\end{equation}

\subsubsection{Flow Predictor Layer} \label{sec:flow_pre}
Some objects are not continuously visible between consecutive frames due to occlusion. The occlusion mask  $O$ is used to distinguish occluded points and non-occluded points in point clouds. 

When estimating the scene flow, to weaken the influence of occluded points, the information of the occlusion mask is included in the formula for calculating flow feature $FF^l$:
\begin{equation}
 FF^l=MLP(F^l \oplus FF_{up}^l\oplus CV_o^l \oplus O^l ), \label{eq:flow_FF}
\end{equation}
where $F^l$ is the feature of point cloud $P^l$. Finally, the formulas  for residual flow $\Delta SF^l$ and scene flow $SF^l$ estimation are:
\begin{equation} 
    \Delta SF^l =FC(FF^l),
\end{equation}
\begin{equation} 
 SF^l = \Delta SF^l + SF^{l}_{up}.
\end{equation}

\section{Loss Function} \label{sec:loss_function}
\subsection{Unsupervised Loss of Scene Flow}
\subsubsection{Chamfer Loss} 
The predicted scene flow $SF^l = \{ sf^l_i |sf_i \in \mathbb{R}^3 \}_{i=1}^{N^{l}}$ at the $l$-th level is added to point cloud $P^l$ of the first frame to get the warped point cloud $P_w^l= \{ p_{w,i}^l | p_{w,i}^l \in \mathbb{R}^3 \}_{i=1}^{N^{l}}$ which is expected as close to the target point cloud $Q^l=\{ q_i^l|q_i^l \in \mathbb{R}^3\}_{i=1}^{N^{l}}$ as possible. The formula is:
\begin{equation} 
\label{equ:2} 
\begin{aligned} 
\ell_C^l = &\sum_{p_w^l \in P_w^l} min_{q_i^l\in Q^l}||p_w^l-q_i^l||_2^2 + \\
  &\sum_{q_i^l \in Q^l} min_{p_w^l \in P_w^l}||p_w^l-q_i^l||_2^2,
\end{aligned} 
\end{equation}
where $||\cdot||_2$ means $l_2$ norm.

\subsubsection{Smoothness Constraint}
To ensure that the scene flow predicted in the local area is smooth, we need to introduce a smooth regularization constraint.  The scene flow of the point $p_i^l$ can be constrained by the scene flow of its $K$ neighbors.  The formula is:
\begin{equation}
\ell_S^l = \sum_{p_i^l \in P^l} \frac{1}{|N(p_i^l)|}\sum_{s_k^l \in N(p_i^l)} ||SF^l(p_i^l)-SF^l(s_k^l)||_2^2, \label{eq:smooth}
\end{equation}
 where $N(p_i^l)$ is $K$ nearest neighbors of $p_i$. $|N(p_i^l)|$ is the the number of points in the local region $N(p_i^l)$.

\subsubsection{Laplacian Regularization}
The shape characteristic of a certain point in a point cloud is represented by a Laplace vector as follows:
\begin{equation}
  v(p_i)=\frac{1}{|N(p_i)|} \sum_{s_k \in N(p_i)}(s_k -p_i), \label{eq:laplacian_vector}
\end{equation}
where the $v(p_i)$ is the Laplacian characteristic of point $p_i$. 

The predicted point cloud $P_w^l$ and the target point cloud $Q^l$ have similar local shape characteristics at corresponding points. The Eq. (\ref{eq:laplacian_vector}) is used to calculate the Laplacian characteristic $v(q_i^l)$ of $q_i^l$ and the Laplacian characteristic $v(p_{w,i}^l)$ of $p_{w,i}^l$. Then, is used to obtain the Laplacian characteristic $v(q_{inter}^l)$ of $q_{inter}^l$ is obtained by the inverse distance-based interpolation from $q_i^l$. The $q_{inter}^l$  is the corresponding point of $p_{w,i}^l$ in $P_w^l$. The loss formula is:
\begin{equation}
\ell_L^l = \sum_{s_k^l \in N(p_i^l)} ||v(p_{w,i}^l)-v(q_{inter}^l)||_2^2. \label{eq:laplacian}
\end{equation}
In summary, the scene flow unsupervised loss formula is:
\begin{equation}
L_{sf} = \sum_{l=0}^3 \alpha_l (\sigma_C \ell_C^l + \sigma_S \ell_S^l + \sigma_L \ell_L^l), \label{eq:loss_flow}
\end{equation}
where $\alpha_l$ denotes the weight for each pyramid level. $\sigma_C$, $\sigma_S$, and $\sigma_L$ denote the weight for these three unsupervised losses, respectively.

\subsection{Supervised Loss of Pose}
The pose estimation at the $l$-th level are quaternion $q^l$ and translation vector $t^l$. $q_{gt}$ and $t_{gt}$ are the ground truth generated from the ground truth of pose transformation matrix. The supervised loss of odometry at the $l$-th level is:
\begin{equation}
\begin{aligned}
\ell_{pose}^l=&\Vert t_{gt}-t^l\Vert exp(-w_x)+w_x+\\ &\Vert q_{gt}-\frac{q^l}{|q^l|}\Vert_2 exp(-w_q)+w_q, \label{eq:loss_pose_l}
\end{aligned}
\end{equation}
where $||\cdot||$ means $l_1$ norm. $w_x$ and $w_q$ are learnable parameters introduced in previous work \cite{li2019net}. The supervised loss of odometry is:
\begin{equation}
L_{pose} = \sum_{l=0}^3 \lambda_l \ell_{pose}^l, \label{eq:loss_pose}
\end{equation}
where $ \lambda_l$ denotes the weight for each pyramid level.
\subsection{Total Training Loss}
The total loss of network training is:
\begin{equation}
L = \mu_{sf} L_{sf}+ \mu_{p} L_{pose}, \label{eq:loss_total}
\end{equation}
where $\mu_{sf}$ and $\mu_{p}$ denote the weights for scene flow unsupervised loss and pose supervised loss.

\begin{table*}[t]
\caption{\textbf{Evaluation Results of 3D Scene Flow on the K-SF-142.}  ICP, FGR, and CPD are geometric methods of non-deep learning. To follow previous works \cite{Pointpwc,Justgowith,with_or_without,Ego-Motion} on unsupervised learning of 3D scene flow, we set up the same evaluation metrics. The best results for each evaluation metric are shown in bold. ``Self" represent self-supervised method. `Self ft" means using unsupervised learning for model fine-tuning. The pose ground truth is not provided in the K-SF-142 dataset, we fine-tune our model without odometry assistance.}
  \label{table:kitti}
  \centering
  \begin{threeparttable}
 		\begin{tabular}{c|c|c|cccc|cc}
			\toprule
			Method & Training Set &  Sup.  & EPE3D($m$) & Acc3DS$\uparrow$  & Acc3DR$\uparrow$ &  Outliers3D$\downarrow$ & EPE2D($px$)$\downarrow$ & Acc2D$\uparrow$ \\ \midrule
			ICP \cite{icp} & No &  No   &   0.5181 & 0.0669 & 0.1667 & 0.8712 & 27.6752 & 0.1056  \\  
			FGR \cite{FGR} & No &  No   &   0.4835 & 0.1331 & 0.2851 & 0.7761 & 18.7464 & 0.2876  \\  
			CPD \cite{CPD} & No &  No   &   0.4144 & 0.2058 & 0.4001 & 0.7146 & 27.0583 & 0.1980  \\ 
			Ego-motion \cite{Ego-Motion} & FT3D &  Self & 0.4154 & 0.2209 & 0.3721 & 0.8096 & 15.0605 & 0.3162 \\
			PointPWC-Net \cite{Pointpwc} & FT3D  &  Self & 0.2549 & 0.2379 & 0.4957 & 0.6863 & 8.9439 & 0.3299 \\
			PointPWC-Net \cite{Pointpwc} & K-OD  &  Self & 0.3712 & 0.1992 & 0.4092 & 0.7406 &  --- & --- \\
			Pontes et al. \cite{with_or_without} & FT3D &   Self  & 0.1690 & 0.2171 & 0.4775 & --- & --- & ---  \\
			Mittal et al. \cite{Justgowith} & FT3D & Self  & \bf0.1220 & 0.2537 & 0.5785 & --- & --- & --- \\
			Ours & K-OD  & Self & 0.1279 & \bf0.3997	& \bf0.6948	& \bf0.4799	& \bf4.0656	& \bf0.6533  \\
			
			\midrule 
			PointPWC-Net\cite{Pointpwc} & FT3D &  Self ft    & 0.1770 & 0.1329 & 0.4215 & 0.272 & --- & ---\\
			JGwF \cite{Justgowith} & FT3D & Self ft  & 0.1260 & 0.3200
			& 0.7364 & --- & --- & --- \\
			SFGAN  \cite{sfgan} & FT3D & Self ft  & 0.0983 & 0.3022 & 0.6823 & 0.5584 & --- & ---\\
			Ours & K-OD & Self ft  &\bf0.0830  &\bf0.5266  &\bf0.8317 &\bf0.3958 &\bf2.7929 &\bf0.7859 \\
			\bottomrule
		\end{tabular}
  \end{threeparttable}
\end{table*}
\begin{table*}[t]
\caption{\textbf{Evaluation Results of 3D Scene Flow on the lidarKITTI}. \cite{flot,Pointpwc,jin2022deformation,Justgowith} do not provide the evaluation results on lidarKITTI. The evaluation results below are obtained by evaluating on lidarKITTI using the pre-trained models they provided. ``Self ft" means using unsupervised learning for model fine-tuning. ``Full ft" means using supervised learning for model fine-tuning.}
  \label{table:lidarKITTI}
  \centering
  \begin{threeparttable}
     \begin{tabular}{c|c|c|cccc}
			\toprule
			Method & Training Set &  Sup.  & EPE3D($m$)$\downarrow$ & Acc3DS$\uparrow$  & Acc3DR$\uparrow$ & Outliers3D$\downarrow$ \\
			\midrule
			FLOT \cite{flot} &FT3D  &Full &0.653 &\bf0.155 &0.313 &0.837  \\
			PointPWC-Net \cite{Pointpwc} & FT3D  &Self & 1.194 &0.038 &0.141 & 0.934 \\
			
			Mittal et al. \cite{Justgowith} & FT3D  &Self & 0.977 &0.010 &0.052 & 0.994 \\

Jin et al. \cite{jin2022deformation} &FT3D &Self &0.590 &0.151 &0.333 &0.849\\

			Ours  & K-OD & Self  &\bf 0.394  &0.116  &\bf0.344 &\bf0.792 \\
			\midrule
			Ours  & K-OD & Self + Self ft  &0.333  &0.139 &0.395 &0.772 \\
			Ours  & K-OD & Self + Full ft  &\bf0.116  &\bf0.413  &\bf0.759 &\bf0.486 \\
			\bottomrule
		\end{tabular}
  \end{threeparttable}
\end{table*}
\section{Experiments} \label{sec:experiment}
\subsection{Implementation Details and Training Details of the Proposed Network} \label{sec: detail}

\subsubsection{Implementation Details} \label{sec: weight}
The loss function of the network consists of two parts, the scene flow unsupervised loss and the pose supervised loss. In the scene flow unsupervised loss, the weights of each layer loss are ${\alpha _0} = 0.02$, ${\alpha _1} = 0.04$, ${\alpha _2} = 0.08$, and ${\alpha _3} = 0.16$.  The weights for three scene flow unsupervised loss are $\sigma_C=1.0$, $\sigma_S=1.0$, and $\sigma_L=0.3$.  In the pose supervised loss, the weights of each layer loss are ${\lambda _0} = 0.2$, ${\lambda _1} = 0.4$, ${\lambda _2} = 0.8$, and ${\lambda _3} = 1.6$. In the total loss, the weights are $\mu_{sf} = 1$ and $\mu_{p} = 1$. We divide the training process into three stages: training only the odometry network until convergence, training only the scene flow network until convergence, and training with both networks together until convergence.

As shown in Fig.  \ref{fig:network}, the number of the points input to the network is $4N = 8192$. The number of downsampling point cloud in each layer are $N^1= N= 2048$, $ N^2 = \frac{1}{2}N =1024 $, $ N^3 = \frac{1}{8}N =256$, and $N^4 = \frac{1}{32}N = 64 $ respectively.

\subsubsection{Training Details}
The environment for the experimental configuration is GeForce RTX 3090 GPU with CUDA = 11.3 and PyTorch = 1.10.0. The optimizer is Adam \cite{kingma2014adam} with ${\beta _1} = 0.9$ and ${\beta _2} = 0.999$.  The initial learning rate is 0.001. The learning rate will decay exponentially with a decay step of 13 epochs and a decay rate of 0.7. The batch size is 8. 

\subsection{Datasets}
\subsubsection{Training Dataset}
FlyingThings3D (FT3D) \cite{Flyingthing3d} dataset  provides more than 20,000 pairs of synthesized point clouds and corresponding scene flow annotations. FT3D is often used to train scene flow estimation models due to its large data with scene flow annotations. However, there are significant differences between the synthesized dataset and the real-world dataset, which results in models trained on the synthesized dataset not being well adapted to the real-world scene. The KITTI Odometry (K-OD) dataset \cite{kitti2013} is obtained by using Velodyne 64-beam LiDAR. It provides 11 sequences, each of which contains LiDAR point clouds and pose ground truth. The pose ground truth is used to supervise the odometry training. Our method learns 3D scene flow on continuous frames of LiDAR point clouds of sequences 00-06 in an unsupervised way. The ground points of each frame of point cloud are removed. 

\subsubsection{Evaluation Dataset}

\begin{itemize}
\item \textbf{K-SF-142}. KITTI Scene Flow 2015 (K-SF)  \cite{Menze2015Joint3E, kitti2018}  consists of 200 training scenes and 200 test scenes which are used for evaluations of RGB stereo-based methods. K-SF-142 consists of 142 pairs in K-SF training scenes. PointPWC-Net \cite{Pointpwc} selected these 142 pairs for evaluation. These point clouds and ground truth flow are generated by projecting annotated disparity maps and optical flow to 3D. We also choose these 142 pairs as our test dataset. The ground points are removed from the point clouds of evaluation datasets like previous methods \cite{Ego-Motion,Pointpwc,Justgowith,with_or_without}. For the fine-tuning experiment, we divide them into 100 training samples and 42 test samples.

\item \textbf{lidarKITTI}. lidarKITTI \cite{geiger2012we} is a real-world dataset collected by using a Velodyne 64-beam LiDAR. There are the same 142 pairs as K-SF-142.  The dataset is obtained by projecting the point cloud onto the image plane and assigning it annotated 3D scene flow.  In the dataset, the points of two input frames do not correspond directly and have a typical LiDAR sensor sampling pattern. For the fine-tuning experiment, we use 100 pairs for fine-tuning training and 42 pairs for testing. 
\end{itemize}

Because the K-SF-142 dataset is synthesised from optical flow and disparity map, the points in the point clouds of two consecutive frames correspond point by point in the K-SF-142 dataset. The previous method selects 8192 points at random in each of the two frames so that the selected points do not correspond point by point. The lidarKITTI is the raw LiDAR point cloud dataset, where the points in the point clouds do not correspond point by point, and the data suffers noise from the LiDAR sensor. Therefore, the lidarKITTI dataset is more challenging.

\begin{figure*}[t]
\centering  
\subfigure[PointPWC-Net (K-OD)]{
\includegraphics[width=0.95\textwidth]{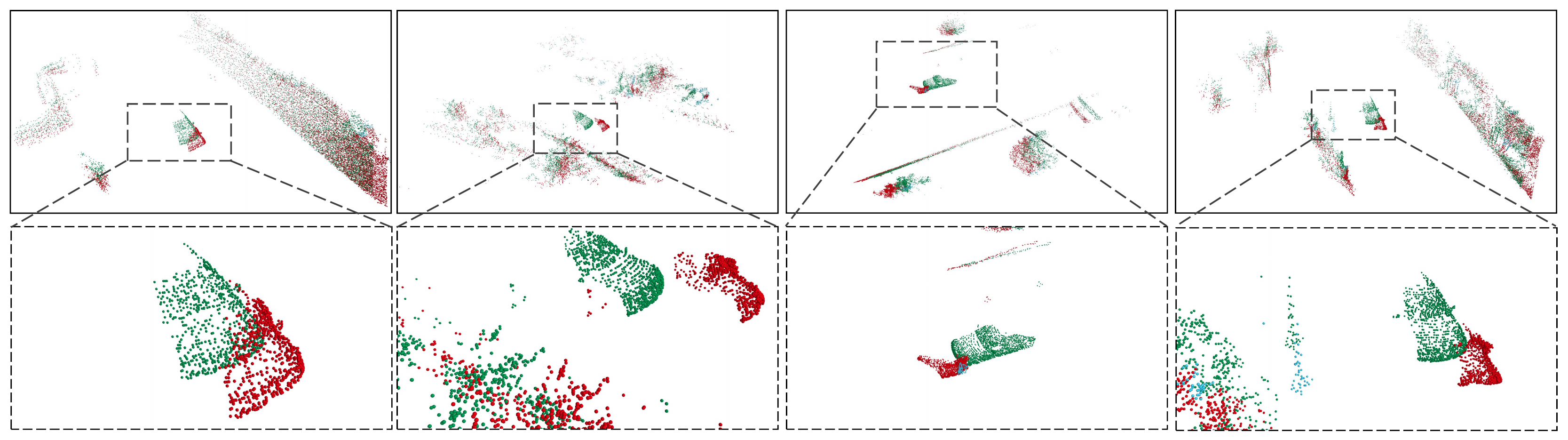}}
\subfigure[Ours (K-OD)]{
\includegraphics[width=0.95\textwidth]{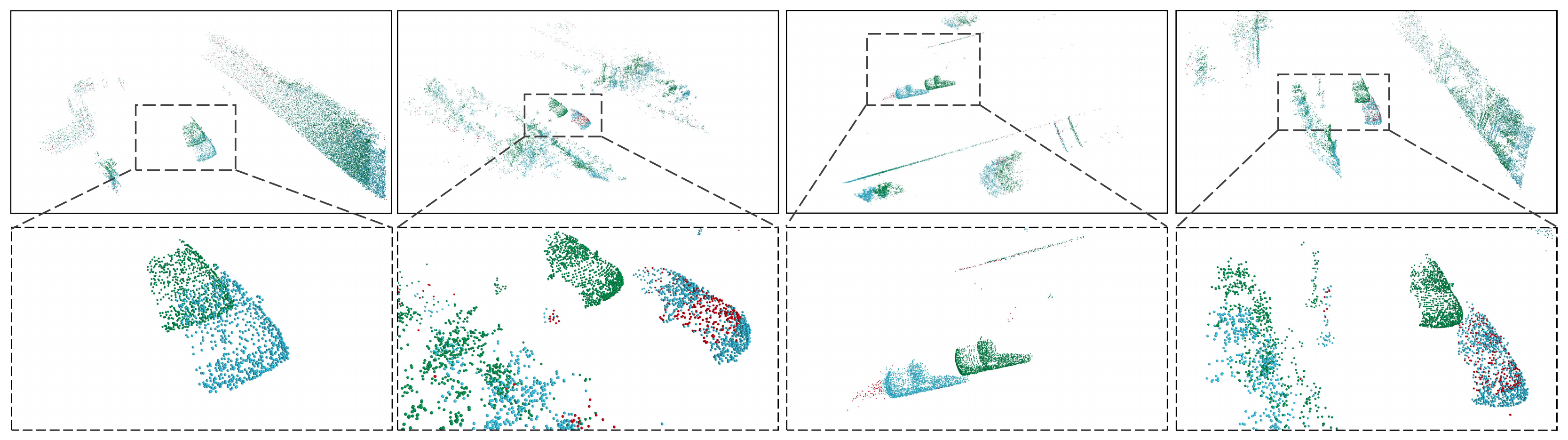}}
\caption{\textbf{Visualization of scene flow estimation results for our method and baseline \cite{Pointpwc}.} The results of PointPWC-Net and our proposed method are based on models trained on the K-OD dataset. The estimated point cloud of the second frame is generated by warping the source point cloud with the predicted scene flow. Green points indicate the source point cloud. The estimated point cloud for the second frame is marked as blue and red. The blue points are considered to be correct  points and the red points are considered to be incorrect points (classified by Acc3D Relax).}
\label{fig:vis0}
\end{figure*}

\begin{figure}[t]
	\centering
	\includegraphics[width=0.48\textwidth]{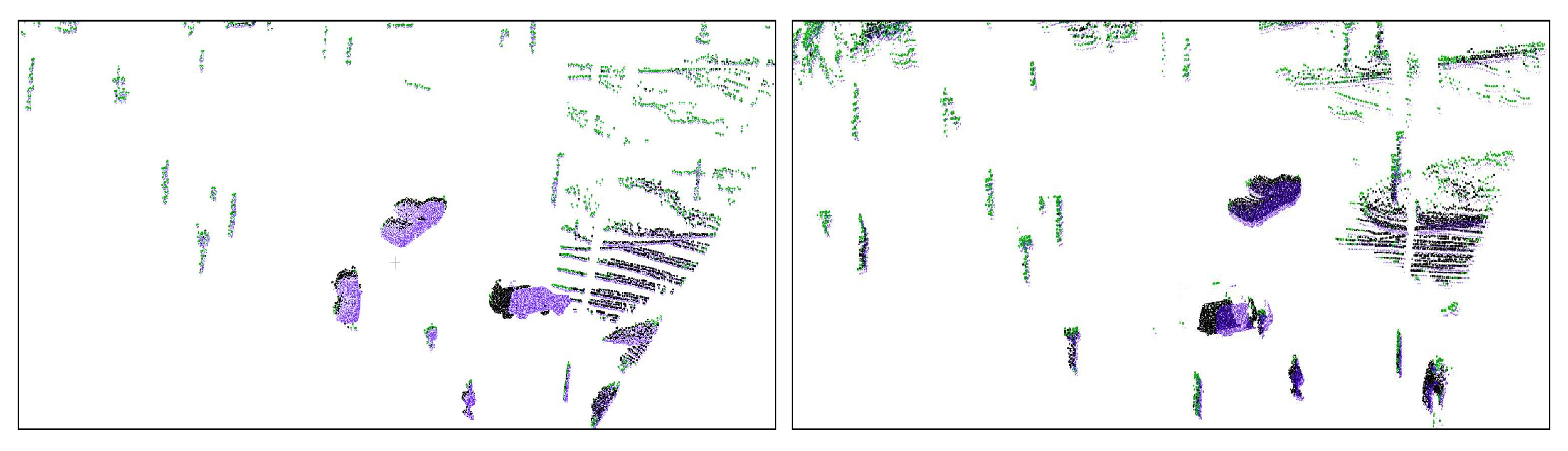}
	\caption{\textbf{Visualization of static mask on lidarKITTI.}
	The point cloud of the first frame is represented in green and black, where the dynamic points are marked in black and the static points are marked in green. The second frame is shown in purple.}
	\label{fig:vis2}
\end{figure}

\begin{figure*}[t]
\centering  
\subfigure[K-SF-142]{
\label{Fig.sub.1}
\includegraphics[width=0.95\textwidth]{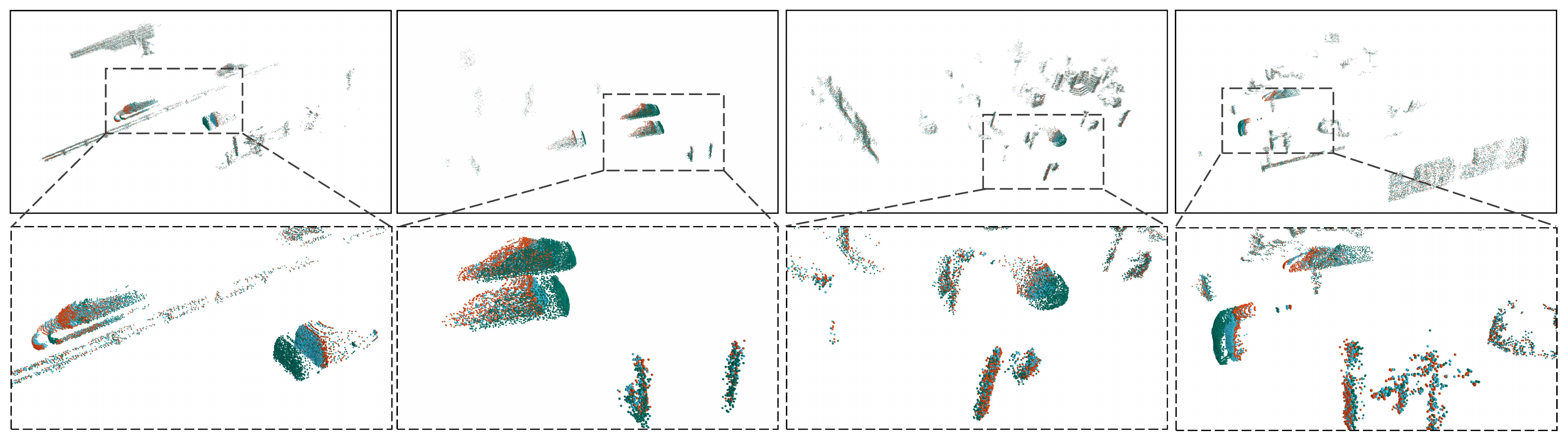}}
\subfigure[lidarKITTI]{
\label{Fig.sub.2}
\includegraphics[width=0.95\textwidth]{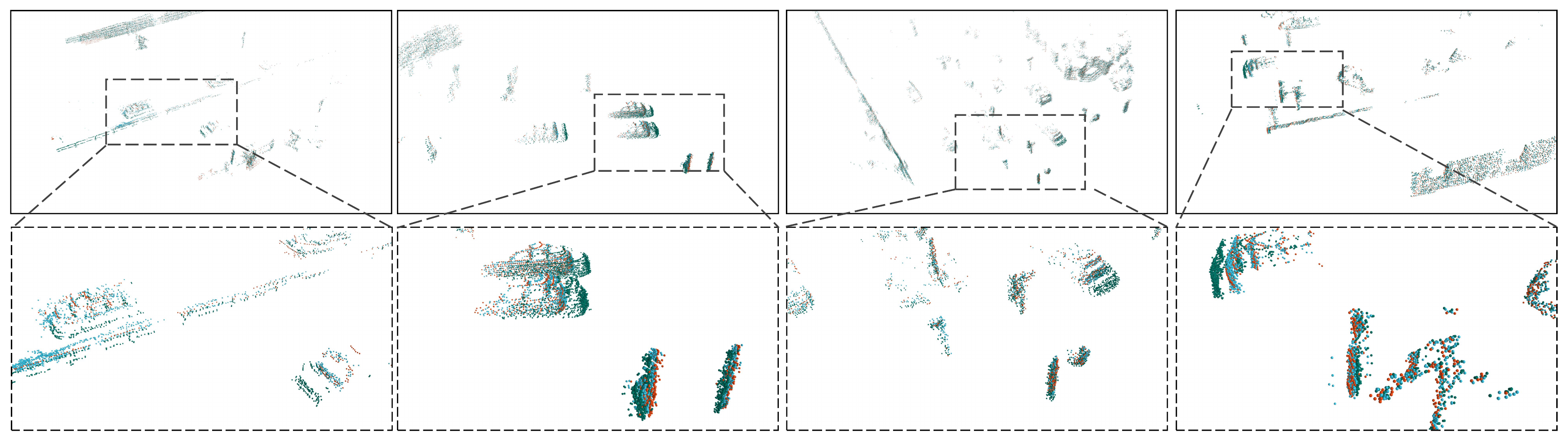}}
\caption{\textbf{Visualization of scene flow estimation and pose estimation on  K-SF-142 and lidarKITTI.} Blue points indicate the estimated point cloud generated by warping the source point cloud with the predicted scene flow. Green points indicate the estimated point cloud generated by warping the source point cloud with the predicted pose. Red points indicate the ground truth of the point cloud in the second frame.}
\label{fig:ksf_lidar}
\end{figure*}

\subsection{Comparison with Other Unsupervised Learning Methods }
In Table  \ref{table:kitti}, the evaluation score of the our proposed model on real-world datasets outperform almost all advanced methods of unsupervised learning \cite{Ego-Motion,Pointpwc,with_or_without,Justgowith}. Compared to the model of PointPWC-Net trained on the FT3D dataset, the experiment results of our method show a substantial improvement in all metrics. Specifically, when PointPWC-Net is trained on the raw LiDAR point cloud, it suffers a certain magnitude of degradation in the performance of scene flow estimation. The sparse nature of the raw LiDAR point cloud weakens its ability to find inter-frame point correspondence. Our model trained on sparse LiDAR point clouds without any fine-tuning still demonstrates competitive results in scene flow estimation. This is attributed to the  odometry-assisted cost volume for 3D scene flow in this paper and the designed  divide-and-conquer warp operation.

In Table \ref{table:lidarKITTI}, the evaluation results of FLOT \cite{flot} are based on the model trained in a supervised way. Our evaluation results are based on the model trained in K-OD in an unsupervised way. The performance of ours without fine-tuning in real-world point clouds markedly outperforms other methods. Because the pose ground truth is not provided in the K-SF-142 dataset, we can only fine-tune our model without the odometry assistance. After fine-tuning, the model performs the best results. The quantitative results demonstrate the powerful learning capability of our proposed network.
\begin{table*}[t]
	\caption{\textbf{Ablation Experiment Results of 3D Scene Flow}}
	\label{table:ablation}
	\centering
	\begin{threeparttable}
		\begin{tabular}{c|cccc|cccccc}
		\toprule
		Dataset & Static mask & Occlusion mask & Pose  & Flow & EPE3D($m$)$\downarrow$ & Acc3DS$\uparrow$  & Acc3DR&  Outliers3D$\downarrow$ & EPE2D($px$)$\downarrow$ & Acc2D$\uparrow$ \\ 
		\midrule
		\multirow{5}*{K-SF-142}
		& & & &\Checkmark &0.1300 &0.3851 &\bf0.6984 &0.4817 &4.1732 &0.6516 \\ 
		& & &\Checkmark & &0.1698 &0.2204 &0.5352 &0.5922 &5.3077 &0.5349\\
		& \Checkmark & &\Checkmark  &\Checkmark &0.1320 &0.3714 &0.6867 & 0.4932 &4.2717 &0.6341 \\
		&  & \Checkmark  &\Checkmark & \Checkmark&0.1350 &0.3674 &0.6720 & 0.4975 &4.2797 &0.6328 \\
		& \Checkmark & \Checkmark  &\Checkmark &\Checkmark & \bf0.1279 & \bf0.3997	& 0.6948	& \bf0.4799	& \bf4.0656	& \bf0.6533 \\
		\midrule 
		\multirow{5}*{lidarKITTI}
		& & & &\Checkmark &0.3959&0.1123&0.3439&0.7963&---&---\\ 
		& & &\Checkmark &&0.4648&0.0329&0.1616&0.8974&---&---\\
		& \Checkmark & &\Checkmark  &\Checkmark&0.4167&0.0925&0.3039&0.8089&---&---\\
		& &\Checkmark &\Checkmark &\Checkmark &0.4053&0.1044&0.3250&0.8079&---&---\\
		& \Checkmark & \Checkmark  &\Checkmark &\Checkmark &\bf0.3943&\bf	0.1159&\bf0.3443&\bf0.7923&---&--- \\
			\bottomrule
		\end{tabular}
	\end{threeparttable}
\end{table*}
\subsection{Visualization of Experimental Results}
The results of PointPWC-Net \cite{Pointpwc} and our proposed method trained on the K-OD dataset are shown in Fig.  \ref{fig:vis0}. Compared with PointPWC-Net, our method correctly estimates the 3D motion of most of the points in the scene without using any scene flow annotations.

The estimated soft static masks are computed as zero-one masks for visualization. The soft static masks with less than 0.40 are identified as dynamic points. As shown in Fig.  \ref{fig:vis2}, our method can clearly distinguish the object that is moving relative to the background points. Although some static points that are closer to the sensor are classified as dynamic points, it does not have a large bad effect on the unsupervised learning of the network. 

The visualization of scene flow estimation and pose estimation is shown in Fig. \ref{fig:ksf_lidar}. For static points, the predicted points obtained by the pose estimation and scene flow estimation are close to the ground truth of the point cloud in the second frame. For dynamic points, the predicted points obtained by the pose estimation have a large error, while the scene flow is able to obtain the correct motion of the points. Since the static mask and occlusion mask distinguish points in different states, the divide-and-conquer point cloud warp layer makes the predicted point cloud more accurate.

\subsection{Ablation Study}

In the point cloud warp layer, to verify the effectiveness of the occlusion mask and static mask, ablation experiments with the occlusion mask or static mask removed are performed. To verify the effectiveness of the scene flow and the pose information, ablation experiments using only the scene flow or only the pose information are performed. The experiment results are shown in Table  \ref{table:ablation}. The best results for each metric are shown in bold. The best results of the evaluation are obtained when all components are used, which demonstrates the effectiveness of the odometry assistance  and the  divide-and-conquer strategy.

\section{Conclusions}\label{sec:conclusion}
In this paper, we propose an unsupervised learning method of scene flow with the assistance of odometry. The odometry information makes up for the deficiencies of the scene flow with a shared cost volume which is trained with the supervised pose loss. In addition, because the accuracy of the scene flow transform and the pose transform is different for the warping of points in different states, we use both occlusion masks and static masks as weights to get a more accurate point cloud  when performing warp operations. It is achieved to compensate the drawback of existing networks to estimate the scene flow of points in different states. Finally, our work demonstrates the feasibility of using a network to learn odometry and scene flow, and can inspire other multi-task learning with relevance.


%
\ifCLASSOPTIONcaptionsoff
  \newpage
\fi

\bibliographystyle{IEEEtran}  
\bibliography{IEEEabrv,bare_jrnl} 
\vspace{-15mm}
\begin{IEEEbiography}[{\includegraphics[width=1in,height=1.25in,clip,keepaspectratio]{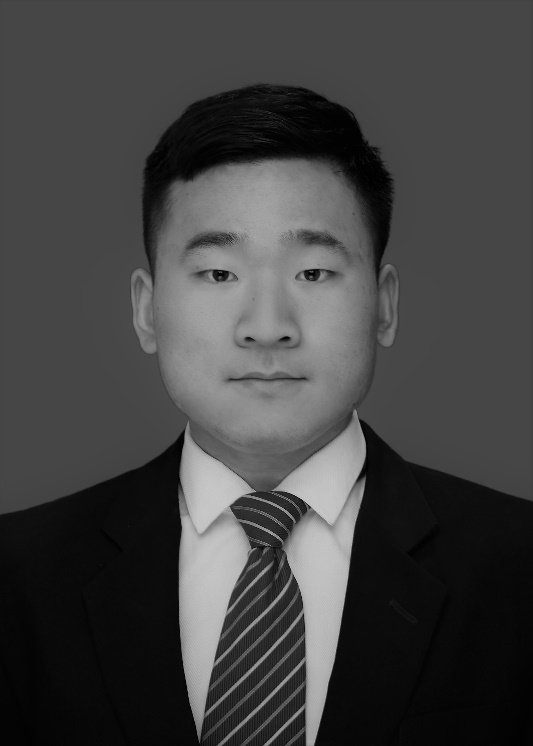}}]{Guangming Wang}
received the B.S. degree from Department of Automation from Central South University, Changsha, China, in 2018. He is currently pursuing the Ph.D. degree in Control Science and Engineering with Shanghai Jiao Tong University. His current research interests include SLAM and computer vision, in particular, 3D scene flow estimation and LiDAR odometry.
\end{IEEEbiography}
\vspace{-180mm}
\begin{IEEEbiography}[{\includegraphics[width=1in,height=1.25in,clip,keepaspectratio]{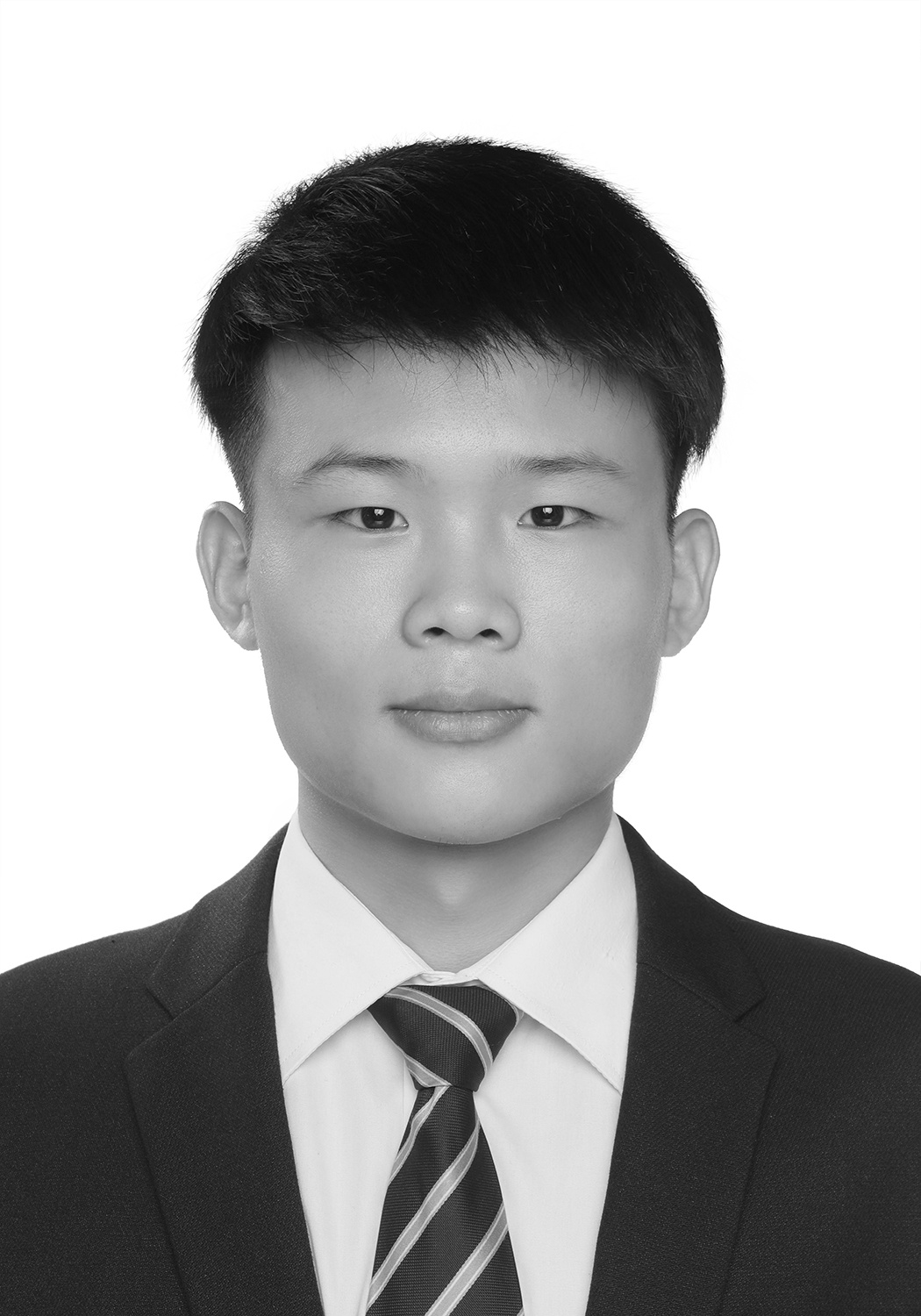}}]{Zhiheng Feng}
received the B.S. degree from Department of Automation from Shanghai Jiao Tong University, Shanghai, China, in 2022.  His current research interests include SLAM and computer vision, in particular, 3D scene flow estimation and LiDAR odometry.
\end{IEEEbiography}
\newpage
\begin{IEEEbiography}[{\includegraphics[width=1in,height=1.25in,clip,keepaspectratio]{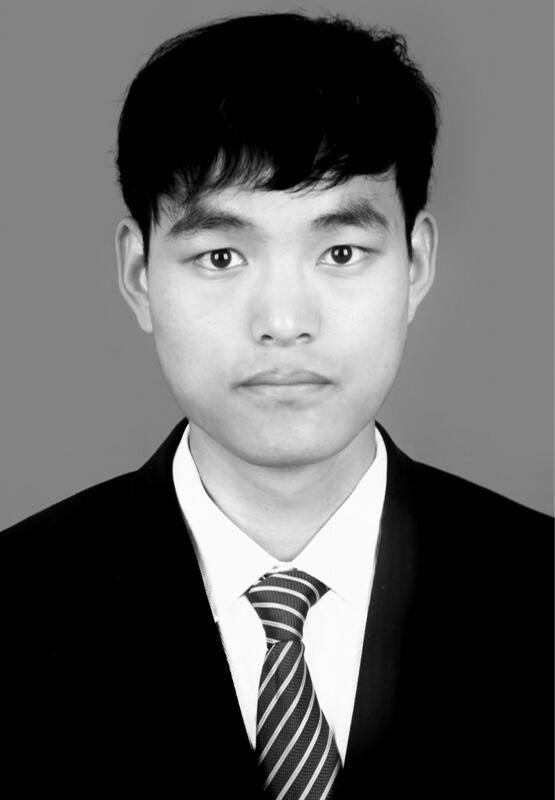}}]{Chaokang Jiang}
received the B.S. degree from the Department of Engineering, Jingdezhen Ceramic Institute College of Technology and Art, Jingdezhen, China, in 2020. He is currently pursuing the M.E. degree in control science and engineering with China University of Mining and Technology. His current research interests include SLAM and computer vision.
\end{IEEEbiography}
\vspace{-150mm}
\begin{IEEEbiography}[{\includegraphics[width=1in,height=1.25in,clip,keepaspectratio]{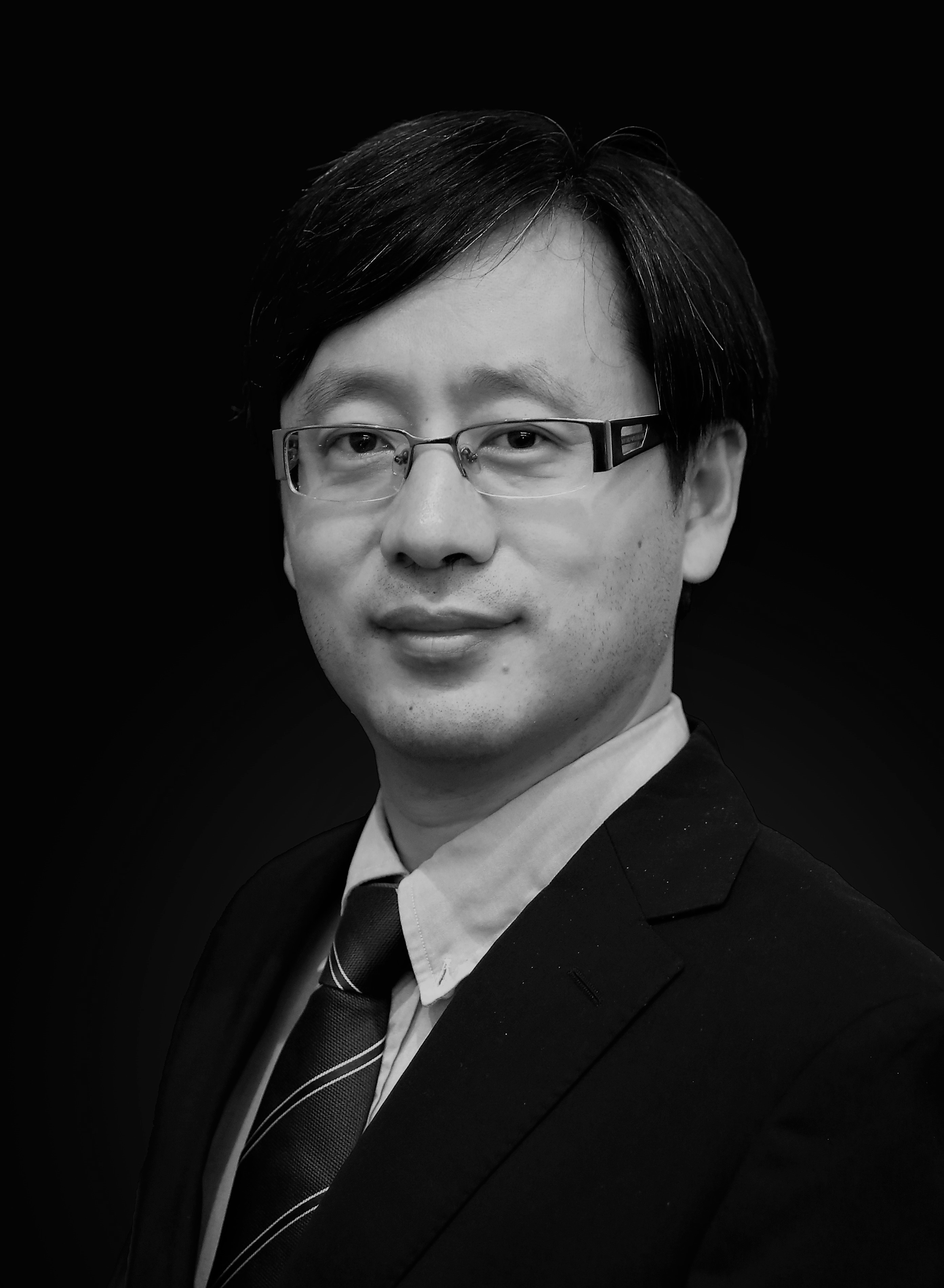}}]{Hesheng Wang} (SM’15) received the B.Eng. degree in electrical engineering from the Harbin Institute of Technology, Harbin, China, in 2002, and the M.Phil. and Ph.D. degrees in automation and computer-aided engineering from The Chinese University of Hong Kong, Hong Kong, in 2004 and 2007, respectively. He is currently a Professor with the Department of Automation, Shanghai Jiao Tong University, Shanghai, China. His current research interests include visual servoing, service robot, computer vision, and autonomous driving. 
Dr. Wang is an Associate Editor of IEEE Transactions on Automation Science and Engineering, IEEE Robotics and Automation Letters, Assembly Automation and the International Journal of Humanoid Robotics, a Technical Editor of the IEEE/ASME Transactions on Mechatronics, an Editor of Conference Editorial Board (CEB) of IEEE Robotics and Automation Society. He served as an Associate Editor of the IEEE Transactions on Robotics from 2015 to 2019. He was the General Chair of IEEE ROBIO 2022 and IEEE RCAR 2016, and the Program Chair of the IEEE ROBIO 2014 and IEEE/ASME AIM 2019. He will be the General Chair of IEEE/RSJ IROS 2025.
\end{IEEEbiography}




\end{document}